\documentclass[11pt,a4paper]{article}
\usepackage[hyperref]{acl2019}
\usepackage{times}
\usepackage{latexsym}
\usepackage{graphicx}
\usepackage{url}

\usepackage{amsmath}

\aclfinalcopy % Uncomment this line for the final submission
\title{Look Again at the Syntax: Relational Graph Convolutional Network \\ for Gendered Ambiguous Pronoun Resolution }

\author{Yinchuan Xu\thanks{* Equal contribution.}\footnote[1]{}\\
  University of Pennsylvania \\ Philadelphia, PA, 19104, USA \\
  \texttt{yinchuan@seas.upenn.edu}
  \\\And
  Junlin Yang\footnote[1]{} \\
  Yale University \\ 
  New Haven, CT, 06511, USA\\
  \texttt{junlin.yang@yale.edu} \\}

\date{}

\begin{document}
\maketitle
\begin{abstract}
Gender bias has been found in existing coreference resolvers. In order to eliminate gender bias, a gender-balanced dataset Gendered Ambiguous Pronouns (GAP) has been released and the best baseline model achieves only 66.9\% F1. Bidirectional Encoder Representations from Transformers (BERT) has broken several NLP task records and can be used on GAP dataset. However, fine-tune BERT on a specific task is computationally expensive. In this paper, we propose an end-to-end resolver by combining pre-trained BERT with Relational Graph Convolutional Network (R-GCN). R-GCN is used for digesting structural syntactic information and learning better task-specific embeddings. Empirical results demonstrate that, under explicit syntactic supervision and without the need to fine tune BERT, R-GCN's embeddings outperform the original BERT embeddings on the coreference task. Our work significantly improves the snippet-context baseline F1 score on GAP dataset from 66.9\% to 80.3\%. We participated in the Gender Bias for Natural Language Processing 2019 shared task, and our codes are available online. \footnote{Our codes and models are available at: \url{https://github.com/ianycxu/RGCN-with-BERT}.}
  
\end{abstract}

\section{Introduction}

Coreference resolution aims to find the linguistic mentions that refer to the same real-world entity in natural language~\cite{pradhan2012conll}. Ambiguous gendered pronoun resolution is a subtask of coreference resolution, where we try to resolve gendered ambiguous pronouns in English such as "he" and "she". This is an important task for natural language understanding and a longstanding challenge.  According to~\cite{sukthanker2018anaphora}, there are two main approaches: heuristics-based approaches and learning-based approaches, such as mention-pair models, mention-ranking models, and clustering models~\cite{mccarthy1995using,haghighi2010coreference,fernandes2014latent}. Learning-based approaches, especially deep-learning-based methods, have shown significant improvement over heuristics-based approaches.

%Among these approaches, Lee et al. ~\shortcite{lee2017end} model shows the effectiveness of neural networks on this task and achieves an overall F1 of 64.0\% on ambiguous gendered pronoun resolution task of  Gendered Ambiguous  Pronouns  (GAP) dataset~\cite{webster2018gap}.

 However, most state-of-art deep-learning-based resolvers utilize one-directional Transformers~\cite{stojanovski2018coreference}, limiting the ability to handle long-range inferences and the use of cataphors. Bidirectional Encoder Representations from Transformers, or BERT~\cite{devlin2018bert} learns a bidirectional contextual embedding and has the potential to overcome these problems using both the previous and next context. However, fine-tuning BERT for a specific task is computationally expensive and time-consuming.

Syntax information has always been a strong tool for semantic tasks. Most heuristics-based methods use syntax-based rules ~\cite{hobbs1978resolving,lappin1994algorithm,haghighi2009simple}. Many of learning based models also rely on syntactic parsing for mention or entity extraction algorithms and compute hand-crafted features as input~\cite{sukthanker2018anaphora}.

Can we learn better word embeddings than BERT on the coreference task with the help of syntactic information and without computationally expensive fine-tuning of BERT? Marcheggiani and Titov et al.~\shortcite{marcheggiani2017encoding} successfully use Graph Convolutional Networks (GCNs)~\cite{duvenaud2015convolutional,kipf2016semi} to learn word embeddings for the semantic role labeling task and outperform the original LSTM contextual embeddings.

Inspired by Marcheggiani and Titov~\shortcite{marcheggiani2017encoding}, we create a 'Look-again' mechanism which combines BERT with Gated Relational Graph Convolutional Networks (R-GCN) by using BERT embeddings as initial hidden states of vertices in R-GCN. R-GCN's structure is derived from a sentence's syntactic dependencies graph. This architecture allows contextual embeddings to be further learned and encoded into better task-specific embeddings without fine tuning BERT which is computationally expensive.

\section{Contributions}
Our main contributions are: (1) Our work is the first successful attempt of using R-GCN to boost the performance of BERT contextual embeddings without the need to fine tune BERT. (2) Our work is the first to use R-GCN on the coreference resolution task. (3) Our work improves the snippet-context baseline F1 score on  Gendered Ambiguous Pronouns dataset from 66.9\%  to  80.3\%.

\section{Methodology}
We propose a series connection architecture of pre-trained BERT with Gated Relational  Graph  Convolutional  Network (Gated R-GCN). Gated R-GCN is used for digesting structural syntactic information. This architecture, which we name as 'Look-again' mechanism can help us learn embeddings which have better performance on coreference task than original BERT embeddings.

\subsection{Syntactic Structure Prior}
 As mentioned in the Introduction section, syntactic information is beneficial to semantic tasks. However, how to encode syntactic information directly into deep learning systems is difficult. 

Marcheggiani and Titov~\shortcite{marcheggiani2017encoding} introduces a way of incorporating syntactic information into sequential neural networks by using GCN. The syntax prior is transferred into a syntactic dependency graph, and GCN is used to digest this graph information. This kind of architecture is utilized to incorporate syntactic structure prior with BERT embeddings for coreference task in our work.

\subsection{GCN}
Graph Convolutional Networks (GCNs) ~\cite{duvenaud2015convolutional,kipf2016semi} take graphs as inputs and conduct convolution on each node over their local graph neighborhoods. The convolution process can also be regarded as a simple differentiable message-passing process. The message here is the hidden state of each node. 

Consider a directed graph $\mathcal{G}= (\mathcal{V},\mathcal{E})$ with nodes $v_i \in \mathcal{V}$ and edges $(v_i, v_j) \in \mathcal{E}$. The original work of GCN~\cite{kipf2016semi} assumes that every node $v$ contains a self-loop edge, which is $(v_i, v_i) \in \mathcal{E}$. We denote hidden state or features of each node $v_i$ as $h_i$, and neighbors of each node as $\mathcal{N}(v_i)$, then for each node $v_i$, the feed-forward processing or message-passing processing then can be written as:

\begin{equation}
h_i^{(l+1)} = ReLU\left(\sum_{u\in \mathcal{N}(v_i)}\frac{1}{c_i} W^{(l)} h_u^{(l)}\right)
\end{equation}

Note that we ignore the bias term here. $l$ here denotes the layer number, and $c_i$ is a normalization constant. We use $c_{i}=|\mathcal{N}(v_i)|$, which is the in-degree of the node. Weight $W^{(l)}$ is shared by all edges in layer $l$.

\subsection{R-GCN}
\begin{figure}
\begin{center}
\includegraphics[scale=0.21]{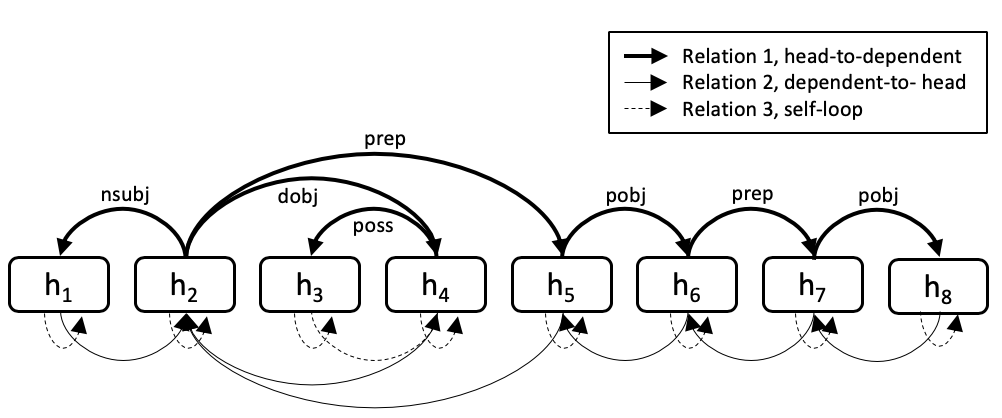}
\end{center}
\caption{Syntactic dependencies graph with three relations} \label{syn_graph}
\end{figure}

Each sentence is parsed into its syntactic dependencies graph and use GCN to digest this structural information. Mentioned in ~\cite{schlichtkrull2018modeling}, when we construct the syntactic graph we also allow the information to flow in the opposite direction of syntactic dependency arcs, which is from dependents to heads.  Therefore, we have three types of edges: first, from heads to dependents; second, from dependents to heads and third, self-loop (see Fig.~\ref{syn_graph}).

Traditional GCN cannot handle this multi-relation graph. Schlichtkrull~\shortcite{schlichtkrull2018modeling} proposed a Relational Graph Convolutional Networks (R-GCNs) structure to solve this multi-relation problem:

\small
\begin{equation}
\begin{split}
h_i^{(l+1)} = ReLU\left( \sum_{r\in R}\sum_{u\in \mathcal{N}_r(v_i)}\frac{1}{c_{i,r}}W_r^{(l)}h_u^{(l)} \right)
\end{split}
\end{equation}
\normalsize

where $\mathcal{N}_r(v_i)$ and $W_r^{(l)}$ denote the set of neighbor of node $i$ and weight under relation $r\in R$ respectively. In our case, we have three relations.

\subsection{Gate Mechanism}
Because the syntax information is predicted by some NLP packages, which might have some error, we need some mechanism to reduce the effect of erroneous dependency edges.

A gate mechanism is introduced in~\cite{marcheggiani2017encoding,dauphin2017language,li2015gated}. The idea is calculating a gate value ranging from $0$ to $1$, and multiplying it with the incoming message. The gate value is computed by:
\begin{equation}g^{(l)}_{u,v} = Sigmoid\left( h_{u}^{(l)} \cdot W_{r,g} \right)
\end{equation}
The final forward process of Gated R-GCN is:
\small
\begin{equation}
\begin{split}
h_i^{(l+1)} = ReLU\left(\sum_{r\in R}\sum_{u\in \mathcal{N}_r(v_i)} g^{(l)}_{u,v_i}\frac{1}{c_{i,r}}W_r^{(l)}h_u^{(l)}\right)
\end{split}
\end{equation}
\normalsize
\subsection{Connect BERT and R-GCN in Series}
We use pre-trained BERT embeddings~\cite{devlin2018bert} as our initial hidden states of vertices in R-GCN. This series connection between pre-trained BERT and Gated R-GCN forms the 'Look-again' mechanism. After pre-trained BERT encodes tokens' embeddings, Gated R-GCN will 'look again' at the syntactic information which is presented as structural information and further learn semantic task-specific embeddings with the explicit syntactic supervision by syntactic structure.

\begin{figure}
\begin{center}
\includegraphics[scale=0.315]{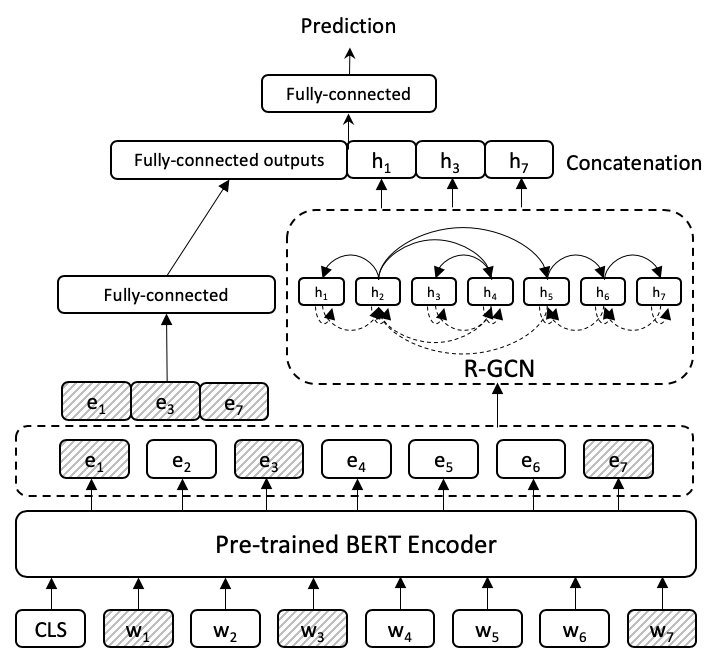}
\end{center}
\caption{End-to-end coreference resolver} \label{nn_archi}
\end{figure}

A fully-connected layer in parallel with Gated R-GCN is utilized to learn a compact representation of BERT embeddings of two mentions (A and B) and the pronoun. This representation is then concatenated with Gated R-GCN's final hidden states of those three tokens. The reason of concatenating R-GCN's hidden states with BERT embeddings' compact representation is that graph convolution of the GCN model is actually a special form of Laplacian smoothing~\cite{li2018deeper}, which might mix the features of vertices and make them less distinguishable. By concatenation, we maintain some original embeddings information. After concatenation, we use a fully-connect layer for the final prediction. The visualization of the final end-to-end model is shown in Fig.~\ref{nn_archi}.

\section{Experimental Methodology and Results}
In the experiment, it shows that, with the explicit syntactic supervision by syntactic structure, Gated R-GCN structure can learn better embeddings that improve performance on the coreference resolution task. Two 
sets of experiments were designed and conducted: Stage one experiments and Full GAP experiments. 

Stage one experiments used the same setting as stage one of shared-task competition, where we had 4454 data samples in total. 'Gap-validation.tsv' and 'gap-test.tsv' were used as training dataset, while 'gap-development.tsv' was used for testing.\footnote{\url{https://www.kaggle.com/c/gendered-pronoun-resolution/data}.}

Full GAP experiments used full 8908 samples of Gendered Ambiguous Pronouns (GAP) dataset in order to compare with the baseline result from the GAP paper~\cite{webster2018gap}.

\subsection{Dataset}
The dataset provided by the shared task is Google AI Language's Gendered Ambiguous Pronouns (GAP) dataset~\cite{webster2018gap}, which is a gender-balanced dataset containing 8,908 coreference-labeled pairs of (ambiguous pronoun, antecedent name), sampled from Wikipedia. 

In stage one of the shared task, only 2454 samples were used as the training dataset, and 2000 samples were used as the test dataset.

\subsection{Data Preprocessing}
SpaCy was used as our syntactic dependency parser. Deep Graph Library (DGL)\footnote{DGL official website: \url{https://www.dgl.ai/pages/about.html}.} was used to transfer each dependency graph into a DGL graph object. Several graphs were grouped together as a larger DGL batch-graph object for batch training setting. R-GCN model was also implemented with DGL.

\subsection{Training settings}
Adam was used~\cite{kingma2014adam} as our optimizer. Learning rate decay was applied. $l2$ regularization of both R-GCN's and fully-connected layer's weights was added to the training loss function. Batch-normalization and drop-out were used in all fully-connection layers. We used one layer for R-GCN which captures immediate syntactic neighbors' information. BERT in our model was not fined tuned and was fixed for training. We used 'bert-large-uncased' version of BERT for generating original embeddings.

The five-fold ensemble was used to achieve better generalization performance and more accurate estimation of the model's performance. The training dataset was divided into 5 folds. Each time of training we trained our model on 4 folds and chose the model which had the best validation performance on the left fold. This best model then was used to predict the test dataset. In the end, predicted results from 5 folds were averaged as the final result.

\subsection{Stage One Experiments}
\begin{figure}
\begin{center}
\includegraphics[scale=0.3]{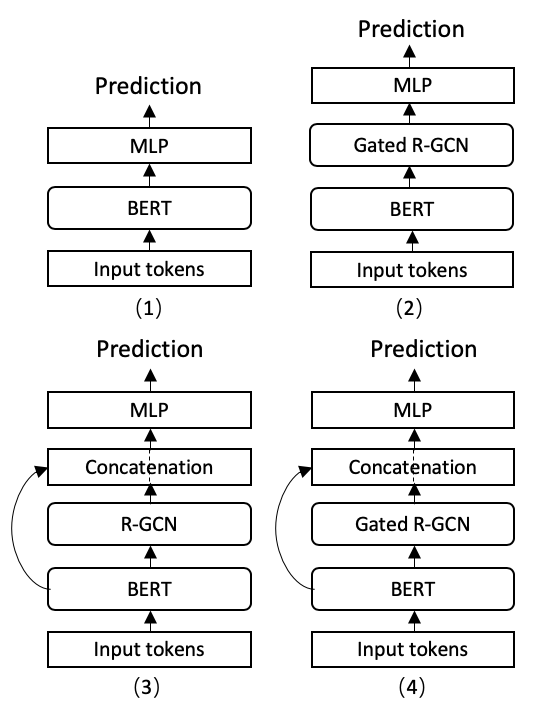}
\end{center}
\caption{ Stage one experiments} \label{experiments}
\end{figure}

There are 4 different settings for Stage One experiments for comparisons (see Fig.~\ref{experiments}):

1. Only BERT embeddings are fed into an additional MLP for prediction. 

2. Connect BERT with Gated R-GCN, but only feed Gated R-GCN's hidden states into MLP for prediction.

3. Connect BERT with R-GCN, and the concatenation is fed into MLP for prediction. The gate mechanism is not applied to R-GCN

4. Connect BERT with Gated R-GCN, and the concatenation is fed into MLP for prediction. The gate mechanism is applied.

\subsubsection{Evaluation Metrics}

The competition used multi-class log-loss as evaluation metrics.

$$log loss = -\frac{1}{N}\sum_{i=1}^N\sum_{j=1}^My_{ij}\log(p_{ij}),$$

where $N$ is the number of samples in the test set, $M$ is 3,  log is the natural logarithm.

\subsubsection{Results}
Table~\ref{stageone} presents the results of four different settings. it demonstrates that R-GCN structure does learn better embeddings and improve the performance. Setting three and setting four show the effectiveness of the Gate Mechanism.

\begin{table}[h]
\begin{center}
\scalebox{0.8}{
\begin{tabular}{|l|l|l|l|l|}
\hline \textbf{BERT} & \textbf{R-GCN}  & \textbf{Concatenation} & \textbf{Gate} & \textbf{Test Log-loss}  \\ \hline 

Yes & No & No & No & 0.5301\\
Yes & Yes & No & Yes & 0.5142\\
Yes & Yes & Yes & No & 0.5045\\
Yes & Yes & Yes & Yes & \textbf{0.4936}\\
\hline
\end{tabular}
}
\end{center}
\caption{\label{stageone} Stage one results }
\end{table}

By comparing setting two and setting four, we can see that because graph convolution of the R-GCN model brings the potential problem of over-smoothing the information~\cite{li2018deeper}, model without concatenation might lose some performance.

\subsection{Full GAP Experiments and Results}
We also tested our model on the full GAP dataset which contains 8,908 samples. 4908 samples were used as training data, and 4000 samples were used as test data. We used micro F1 score as our metric.

The GAP paper~\cite{webster2018gap} introduced several baseline methods: (1) Off-the-shelf resolvers including a rule-based system of Lee et al.~\shortcite{lee2013deterministic} and three neural resolvers from Clark and Manning~\shortcite{clark2015entity}, Wiseman et al.~\shortcite{wiseman2016learning}, and Lee et al.~\shortcite{lee2017end}; (2) Baselines based on traditional cues for coreference; (3)Baselines based on structural cues: syntactic distance and Parallelism; (4) Baselines based on Wikipedia cues; (5) Transformer models~\cite{vaswani2017attention}.

\begin{table}[h]
\begin{center}
\scalebox{1}{
\begin{tabular}{|l|l|}
\hline \textbf{Model}  & \textbf{F1 Score}  \\ \hline 
Lee et al.~\shortcite{lee2017end} & 64.0\%\\
Parallelism & 66.9\%\\
Parallelism+URL  & 70.6\%\\
\hline
BERT only & 78.5\%\\
\textbf{Ours}  & \textbf{80.3\%}\\
\hline
\end{tabular}
}

\end{center}
\caption{\label{font-table} GAP experiments results }
\end{table}

Three best models (Lee et al.~\shortcite{lee2017end}, Parallelism, and Parallelism+URL) from above baselines were chosen for comparison. We first used pre-trained BERT embeddings and fully-connected layers for prediction (see Fig.~\ref{experiments} (1)). Not surprising, BERT embeddings outperformed all of the previous work. 
 
 We then tested our Gated R-GCN model. The model further improved the F1 score by using explicitly syntactic information and learning coreference-task-specific word representations. The final model largely increased the baseline F1 score from 70.6 \% to 80.3 \% and the BERT embeddings' result from 78.5 \% to 80.3 \%.

\subsection{Final Submission and Shared-Task Score}
For the final submission for stage 2 of the shared task, we averaged our result with a BERT-score-layer~\cite{zhang2018neural,clark2016deep} result. In stage two, our work reaches log-loss of 0.394 on the private leaderboard showing that our model is quite effective and robust. This result is obtained without any data augmentation prepossessing.

\section{Discussion and Conclusion}
The Gender Bias for Natural Language Processing (GeBNLP) 2019 shared-task is a competition for building a coreference resolution system on GAP dataset. We participate in this shared-task by using a novel approach which combines Gated R-GCN with BERT. R-GCN is used for digesting syntactic dependency graph and leveraging this syntactic information to help our semantic task. Experiments with four settings were conducted on the shared task's stage one data. We also tested our model on the full GAP dataset where our model improved the best snippet-context baseline F1 score from 66.9 \% to 80.3 \% (by 20 \%). The results showed that, under explicit syntactic supervision and without the need to fine tune BERT,  our gated R-GCN model can incorporate syntactic structure prior with BERT embeddings to improve the performance on the coreference task.

\bibliography{acl2019}
\bibliographystyle{acl_natbib}
\end{document}